\documentclass[journal,12pt,twocolumns]{IEEEtran}
\usepackage{subfig}
\usepackage{graphicx} 
\usepackage{amsmath}

\usepackage{algorithm}
\usepackage{algpseudocode}
\usepackage{float,amsthm}
\usepackage{amsfonts}
\usepackage{booktabs}
\usepackage{multirow}
\usepackage[table]{xcolor}
\usepackage{mathtools}

\theoremstyle{definition}

\newcounter{remark}
\newenvironment{remark}[1][Remark]{\refstepcounter{remark}\begin{trivlist}
\item[\hskip \labelsep {\bfseries #1 \theremark.}]}{\end{trivlist}}

\newcounter{example}
\newenvironment{example}[1][Example]{\refstepcounter{example}\begin{trivlist}
\item[\hskip \labelsep {\bfseries #1 \theexample.}]}{\end{trivlist}}

\newcounter{property}

\def \h {\mathbf h}

\def \y {\mathbf{y}}

\def \w {\mathbf{w}}

\def \I {\mathcal{I}}

\def \z{\mathbf {z}}

\usepackage{color}
\newcommand{\ParSection}[1]{}

\begin{document}
\twocolumn[
\begin{@twocolumnfalse}
\vspace*{5cm}

\copyright 2019 IEEE.  Personal use of this material is permitted.  Permission from IEEE must be obtained for all other uses, in any current or future media, including reprinting/republishing this material for advertising or promotional purposes, creating new collective works, for resale or redistribution to servers or lists, or reuse of any copyrighted component of this work in other works.
\end{@twocolumnfalse}
]
\newpage

\title{Enabling Explainable Fusion in Deep Learning with Fuzzy Integral Neural Networks}

\author{Muhammad~Aminul~Islam,~\IEEEmembership{Member,~IEEE,}
      Derek~T.~Anderson,~\IEEEmembership{Senior~Member,~IEEE,}
       Anthony~J.~Pinar,~\IEEEmembership{Member,~IEEE,}       
       Timothy~C.~Havens,~\IEEEmembership{Senior~Member,~IEEE,}
       Grant~Scott,~\IEEEmembership{Senior~Member,~IEEE,}
       and~James~M.~Keller,~\IEEEmembership{Life~Fellow,~IEEE}
\thanks{Muhammad Aminul Islam is with the Department
of Electrical and Computer Engineering, Mississippi State University, MS, USA e-mail: (mi160@msstate.edu).}
\thanks{Derek Anderson, Grant Scott, and James Keller are with the Electrical Engineering and Computer Science Department, University of Missouri, MO, USA. Timothy Havens and Anthony Pinar are with the Electrical and Computer Engineering Department and the Computer Science Department, Michigan Technological University, MI, USA.}
\thanks{Manuscript received May 20, 2018.}}

\markboth{Author copy, to appear in Special Issue on Deep Fuzzy Models, Transactions on Fuzzy Systems}%
{Shell \MakeLowercase{\textit{et al.}}: Draft Submitted to the IEEE Transactions on Fuzzy Systems}


\maketitle
\begin{abstract}
Information fusion is an essential part of numerous engineering systems and biological functions, e.g., human cognition. Fusion occurs at many levels, ranging from the low-level combination of signals to the high-level aggregation of heterogeneous decision-making processes. While the last decade has witnessed an explosion of research in deep learning, fusion in neural networks has not observed the same revolution. Specifically, most neural fusion approaches are ad hoc, are not understood, are distributed versus localized, and/or explainability is low (if present at all). Herein, we prove that the fuzzy \emph{Choquet integral} (ChI), a powerful nonlinear aggregation function, can be represented as a multi-layer network, referred to hereafter as ChIMP. We also put forth an \emph{improved ChIMP} (iChIMP) that leads to a stochastic gradient descent-based optimization in light of the exponential number of ChI inequality constraints. An additional benefit of ChIMP/iChIMP is that it enables \emph{eXplainable AI} (XAI). Synthetic validation experiments are provided and iChIMP is applied to the fusion of a set of heterogeneous architecture deep models in remote sensing. We show an improvement in model accuracy and our previously established XAI indices shed light on the quality of our data, model, and its decisions.
\end{abstract}

\begin{IEEEkeywords}
data fusion, Choquet integral, deep learning, neural network, explainable AI
\end{IEEEkeywords}

\IEEEpeerreviewmaketitle

\section{Introduction}
Data are ubiquitous in today's technological era. This is both a blessing and a curse as we are swimming in sensors but drowning in data. In order to cope with these data, many systems employ data/information fusion. For example, you are right now combining multiple sources of data, e.g., taste, smell, touch, vision, hearing, memories, etc. In remote sensing, it is common practice to combine lidar, hyperspectral, visible, radar and/or other variable spectral-spatial-temporal resolution sensors to detect objects, perform earth observations, etc. This is the same story for computer vision, smart cars, Big Data, and numerous other thrusts. While the last decade has seen great strides in topics like deep learning, the reality is that our understanding of fusion in the context of \emph{neural networks} (NNs) (and therefore deep learning) has not witnessed similar growth. Most approaches to fusion in NNs are ad hoc (specialized for a particular application) and/or they are not well understood nor explainable (i.e., how are the data being combined and why should we trust system outputs). 

Let $\z_i \in \Re^{D_i}$ be data from source $i = 1, \dots, N$ (sensor, algorithm, human). If fusion is needed, most approaches just concatenate, i.e., $\z = (\z_1,...,\z_N)^t$, resulting in a higher dimensional input. As such, ``fusion'' occurs somewhere in the network. Another approach is divide-and-conquer, where individuals NNs are attached to each $\z_i$, followed by NN(s) (or other machine learning algorithms like a support vector machine) to combine their output. However, whereas this approach gives rise to a modular design, plug-and-play possibilities, etc., it does so at the expense of likely not exposing low-level data correlations (which are in $\z$). In \cite{8281516}, Xiaowei et al.~explored infinite-valued logic on a set of pre-trained \emph{convolutional neural networks} (CNNs). Pal, Mitra, and others (e.g., Keller and the fuzzy perceptron \cite{keller1985incorporating}) explored a variety of topics like fuzzy min-max networks, fuzzy \textit{multilayer perceptron} (MLP), Sugeno fuzzy measure densities \cite{karczmarek2017developing},  and fuzzy Kohonen networks. In 1992 \cite{yager1992applications}, Yager put forth the \emph{ordered weighted average} (OWA) \cite{yager1988ordered}  neuron---which technically is a \emph{linear order statistic} (LOS) as the weights are real-valued numbers (vs. sets). In 1995, Sung-Bae utilized the OWA for NN aggregation at the decision/output level \cite{Sung1995}. 

In 1995, Sung-Bae et al.~explored the fuzzy integral, specifically the Sugeno integral, for NN aggregation \cite{cho1995combining}. They used the \emph{Sugeno $\lambda$-fuzzy measure} ($\lambda$-FM) defined on the $N$ singletons versus the full set of $2^N$ subsets and the densities were derived using their respective accuracy rates on training data. In 2017 \cite{scott2017fusion}, we used the Sugeno and \emph{Choquet integral} (ChI) for \emph{deep CNN} (DCNN) fusion. Specifically, we used data augmentation and transfer learning to adapt GoogLeNet, AlexNet, and ResNet50 from camera imagery to remote sensing imagery. We then applied different aggregations---fuzzy integral, voting, arrogance, and weighted sum---to these DCNNs. A $\lambda$-FM with normalized classifier accuracy densities and also a genetic algorithm was used to learn the densities. In \cite{andersonbryce}, quadratic programming was used to learn the full ChI, relative to pre-trained DCNNs. These are a few NN fusion approaches explored to date.

\ParSection{Why Care?} Herein, we investigate basic NN fusion questions. The first, Q1, is what fusions---aggregation functions to be precise---are possible relative to existing NN ingredients? Q1 is more-or-less an existence argument. The next, Q2, is can we represent and optimize an aggregation function, such as the ChI, as an NN? As such, Q2 addresses how do we find a solution (versus does one exist). Last, Q3, is can we open the hood on a fusion NN and understand what it has learned? 

\ParSection{Novelty and Contributions} The following contributions are made in this paper. For Q1, we demonstrate that state-of-the-art aggregation operators are achievable using existing NN mathematics. Namely, we show that two NNs can compute the ChI; one based on a selection network and $N!$ \emph{linear convex sums} (LCS), the other based on the Mobius transform. We also logically and empirically show that it is a feat to approximate the ChI on limited variety and volume data (which is often the case). For Q2, we present the \emph{ChI multi-layer perception} (ChIMP) (aka, dedicated fusion network) that can be optimized via \emph{stochastic gradient descent} (SGD) in light of an exponential number of ChI inequality constraints. 
For Q3, we use indices for introspection on ChIMP. Whereas most NN fusion solutions to date operate on the basis of implicit and distributed computation, we focus on explicit and centralized computation to promote understandability. ChIMP is used here to fuse a set of heterogeneous architecture deep NNs for remote sensing. Adding ChIMP to the top of a collection of deep NNs results in a deep fuzzy NN.



\ParSection{Organization of Paper} The remainder of this article is organized as such. First, in Section \ref{sec:background} we introduce the capacity and integral. In Section \ref{sec:ChIMP} different NNs (ChIMPs) for the ChI are put forth; Section \ref{sec:ichimp} is an \emph{improved ChIMP} (iChIMP) (relative to SGD optimization) and Section \ref{sec:inds} presents \emph{eXplainable AI} (XAI) fusion. The final sections present our experiments, results and, conclusions.  


\section{Background: Measure and Integral}\label{sec:background}
Let $X=\{x_1,x_2,...,x_N\}$ be $N$ sources of data/information (e.g., sensor, human, algorithm), let $h(x_i)$ be the input from source $i$, and let $\h=(h(x_1),...,h(x_N))^t$ be a vector of inputs. An aggregation operator maps data-to-data, $f_{\Theta}(\h)=\y$, which ideally obeys conditions such as idempotency, associativity, continuity, symmetry, etc. Typically, $\y$ is not multi-dimensional, but is $\Re$-valued. The ChI is a nonlinear aggregation function parameterized by the FM \cite{SugenoPHD,choquet1954theory}. Whereas the integral has its roots in calculus, Keller et al.~were the first to use it for pattern recognition/machine learning \cite{KellerFI,grabisch1992multi,grabisch1994classification}. However, the integral has been used in many contexts, e.g., by Grabisch et al.~in \emph{multi-criteria decision making} (MCDM) \cite{grabisch1996application}. Regardless of the application the question remains: where do the parameters come from? Examples include human specification (which becomes quickly intractable; as $N$ grows, there are $2^N$ variables and $N(2^{N-1})$ monotonicity constraints), it can be learned from training data \cite{islam2017data}, or extrapolated in a crowd sourced fashion \cite{6987326,6251281}. 

In addition, it is important to remark about the complexity of the ChI. For example, for $N=10$ there are $1{,}024$ variables and $5{,}120$ constraints. In order to combat the computational complexity, imputation methods like the $\lambda$-FM have been put forth, where one specifies the measure of the $N$ individuals and the $\lambda$-FM automatically fills in (guesses at) the remaining $N(2^{N-1})-N$ values. Grabish, Labreuche, and others have explored routes like the $k$-additive integral to restrict the number of FM variables to at most $k$ tuples \cite{GRABISCH1997167}. This helps control the complexity of the integral relative to tasks like human decision making and bounded rationality. In \cite{islam2017data}, we introduced a way to identify data supported and data unsupported ChI variables. Optimization is for data supported variables only, new examples are classified as known or unsupported (i.e., requires unknown variables), and imputation is used to make an intelligent guess in the case of missing variables. The next few subsections are quick reviews of the measure and integral.

\subsection{Fuzzy Measure}
The FM, $g:2^X \rightarrow \mathbb{R}^+$, is a function with the following two properties; (i) (boundary condition) $g(\emptyset) = 0$, and (ii) (monotonicity) if $A,B \subseteq X$, and $A \subseteq B$, then $g(A) \le g(B)$.\footnote{Sometimes a normality condition is imposed such that $g(X)=1$.}

\subsection{Choquet Integral}
The ChI of observation $\mathbf{h}$ on $X$ is 
\begin{align}\label{eq:ChI}    
\int{\h \circ g} = C_g(\h) = \sum_{j=1}^N h_{\pi(j)} (g(A_{\pi(j)}) - g(A_{\pi(j-1)})),
\end{align}   
for $A_{\pi(j)} = \{ x_{\pi(1)},$  $\dots,$ $ x_{\pi(j)}\}$, $g(A_{\pi(0)})=0$, and permutation $\pi$ such that 
$h_{\pi(1)}$ $ \ge h_{\pi(2)} $ $ \ge \dots $ $\ge h_{\pi(N)}$.\footnote{Shorthand notation $h_i=h(x_i)$ is used.}

\subsection{Choquet Integral as $N!$ Linear Convex Sum Operators}
One way to discuss the ChI is in terms of $N!$ LCS operators. Relative to a particular sorting of the data ($\pi_i$)---of which there are $N!$ possible sorts---the ChI can be expressed as
\begin{align}
f_{\pi_i} = \sum_{j=1}^N h_{\pi_i(j)} (g(A_{\pi_i(j)}) - g(A_{\pi_i(j-1)})) = \h_{\pi_i}^t\w_{\pi_i},
\end{align}
where $w_{\pi_i}(j)=(g(A_{\pi_i(j)}) - g(A_{\pi_i(j-1)}))$. For the ChI, these $N \times N!$ weights are tied to the underlying $2^N$ FM variables.

\begin{example}
For $N=2$, the ChI can be expanded as
\begin{align}
C_g(\h) = \left\{
  \begin{array}{lr}
    h_1 w_1 + h_2 w_2 & : h_1 \geq h_2 \\
    h_2 w_3 + h_1 w_4 & : h_2 > h_1
  \end{array}
\right.
\end{align}
where the weights are $w_1 = g(\{x_1\})$, $w_2 = 1 - g(\{x_1\})$, $w_3 = g(\{x_2\})$, $w_4 = 1 - g(\{x_2\})$. Thus, there are four weights, but just two underlying free FM variables: the densities.
\end{example}

\begin{remark}
This difference in weights versus underlying FM variables grows with respect to $N$. For example, when $N=5$, there are $32$ FM variables and $600$ weights. However, for $N=10$ there are $1,024$ FM variables and $36,288,000$ weights. The ChI can be seen as a form of variable compression.
\end{remark}

\subsection{Restricting the Scope of the FM/ChI}
Since the ChI is a parametric function, once the FM is determined, the ChI turns into a specific operator. For example: if $g(A)=1, \forall A \in 2^X \setminus \emptyset$, the ChI becomes the maximum operator; if $g(A)=0, \forall A \in 2^X \setminus X$, we recover the minimum; if $g(A)=\frac{|A|}{N}$, we recover the mean; and for $g(A)=g(B)$ when $|A|=|B|$, $\forall A,B \subseteq X$, we obtain an LOS. In general, each of these cases can be viewed as constraints or simplifications on the FM, and therefore the ChI. Also, the reader should know that the all-too-familiar operators---mean, max, min, trimmed versions of these operators, etc.---are all subsets of the LOS and therefore of the ChI. As such, the ChI is useful because it can be adapted to suit a wide range of aggregation needs. This is a primary reason for selecting it for use in this article and for sensor fusion, in general.

\section{The Choquet Integral as an NN: ChIMP}\label{sec:ChIMP}

\begin{figure} 
  \centering
    \includegraphics[width=0.48\textwidth]{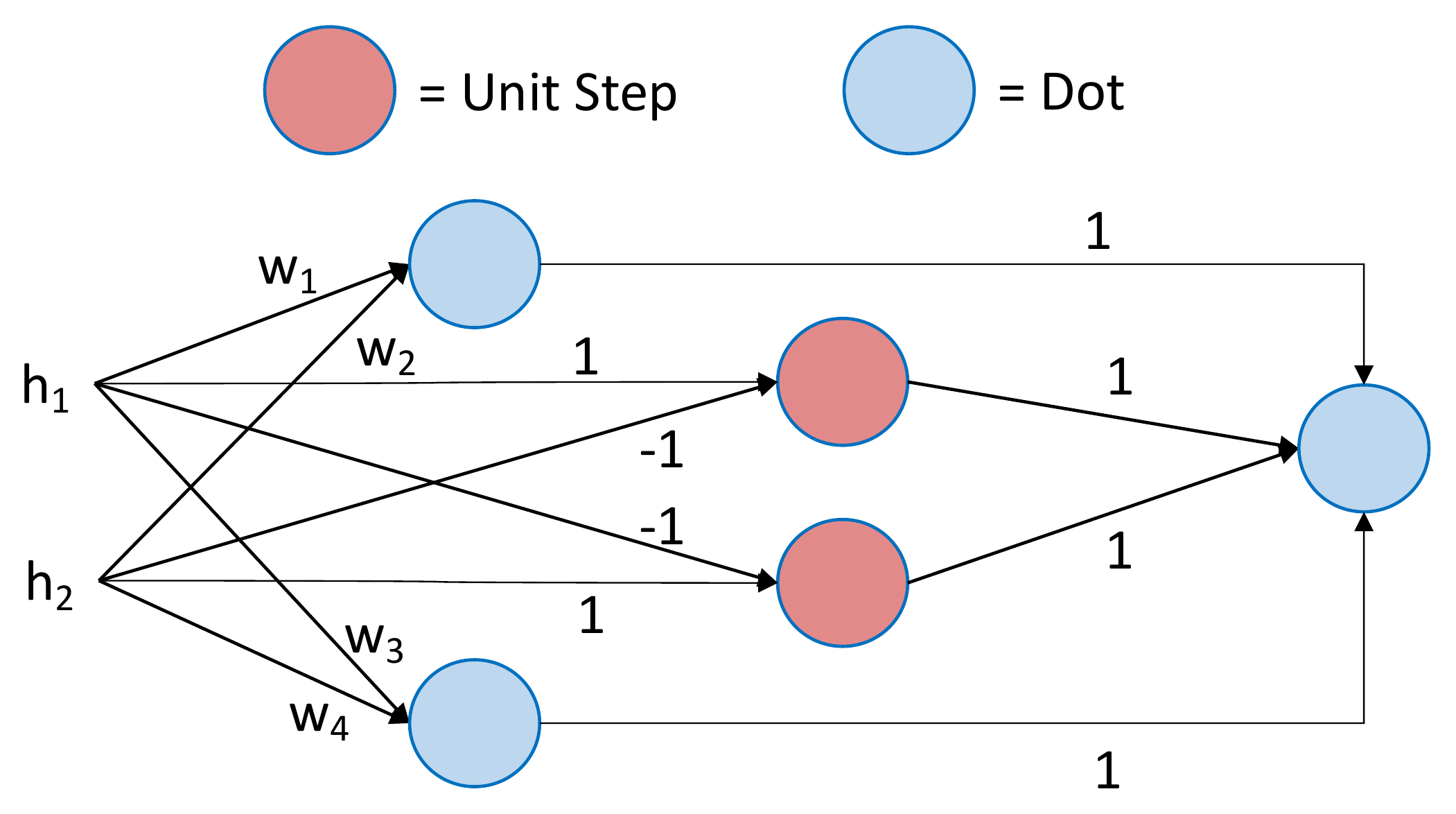}
  \caption{NN to compute the ChI for $N=2$. The first two blue neurons (dot products) on the left are $N!$ LCSs, the red neurons (nonlinearlities) select a LCS (based on input sort order), and the right blue neuron sums the results. Dot is the dot product and Unit Step specifics are outlined in Section \ref{sec:ChIMP}. Multiple inputs to nodes are summed.}\label{fig:chi_mlp}
\end{figure}

In this section we explore question Q1, can an NN compute the ChI, and therefore a wide set of interesting, useful, and commonly used aggregation operators? To make a long story short, the answer is yes (see Example \ref{ex:chimlp} and Fig.~\ref{fig:chi_mlp}). But why is Q1 important? Well, there are many claims about what an NN (e.g., CNN) can do.\footnote{Keeping in mind the difference between an existence proof versus can we identify a way to achieve (e.g., learn) it.} Mathematically, a CNN can encode filters (linear time invariant filters such as a matched filter, low pass filter, or Gabor filter), random projections, and combinations thereof, to name a few. However, limited attention has been placed on understanding aggregation in an NN. Ideally, we would like to know if, and then where, fusion is occurring, understand what aggregation operator was selected (e.g., intersection like, union like, average like, something more exotic), what aggregation functions are possible relative to a network, etc. Understanding if an NN can compute the ChI gives us insight into what is possible, or not possible as it may be.

\begin{example}\label{ex:chimlp}
Consider the case of $N=2$ and the NN outlined in Fig.~\ref{fig:chi_mlp}. The network output is
\[o = u(h_1-h_2)(h_1 w_1 + h_2 w_2) + u(h_2-h_1)(h_2 w_3 + h_1 w_4),\]
where $u$ is a unit/Heaviside step function,\footnote{The unit step function is $u(x)=1$ if $x \geq 0$, otherwise $u(x)=0$. In practice, many use a smooth approximation like the logistic function $\frac{1}{2}+\frac{1}{2}tanh(kx) = \frac{1}{1+e^{-2kx}}$, where the larger the $k$, the sharper the transition about $x=0$.}$^,$\footnote{Herein, we consider a modified unit step function that has value $\frac{1}{N}$ if $f(x) = 0$. In the difference-in-$g$ form of the ChI, what is the rule for the case of equal input values? For example, let $\h=(0.2,0.2,0.1)$. For $\h$ we can choose $\pi(1)=1$, $\pi(2)=2$, and $\pi(3)=3$ or $\pi(1)=2$, $\pi(2)=1$, and $\pi(3)=3$. Depending on the underlying FM, these sorts can yield different ChI outputs. In practice, most sort algorithms use an increasing or decreasing rule (i.e., default mapping to one case). Herein, we augment the unit step function and consider the average function value.} which gives us
\begin{align}
 o &= u(h_1-h_2)\left[h_1 g(\{x_1\}) + h_2 (1-g(\{x_1\}))\right] \nonumber \\
      & + u(h_2-h_1)[h_2 g(\{x_2\}) + h_1 (1-g(\{x_2\}))]; \nonumber
\end{align}
thus,
\[
o  =
\begin{cases}
h_1 g(\{x_1\}) + h_2 (1-g(\{x_1\})), & h_1 > h_2,\\
h_2 g(\{x_2\}) + h_1 (1-g(\{x_2\})), & h_2 > h_1,\\ 
0.5(h_1 g(\{x_1\}) + h_2 (1-g(\{x_1\})))+\\
0.5(h_2 g(\{x_2\}) + h_1 (1-g(\{x_2\}))), & h_1=h_2.
\end{cases}
\]
Without loss of generality, this extends to any $N$.
\end{example}

As the reader can see, ChIMP represents the ChI by a set of LCS operators and it uses a \emph{selection network} to pick one of these results. Technically, our solution can learn and compute the ChI, but it has more functionality (freedom) than a standard ChI as we made the $N! \times N$ weights independent (and in $\Re$ versus $\Re^+$), versus reducing them (sharing weights) into the underlying $2^N$ FM variables, which can be done. However, our goal in this section is not to make the simplest possible network, it is to show that an NN can represent the ChI.

\begin{remark}
As discussed in the introduction, answers for fusion are in the eye of the beholder; that is, context matters. Figure \ref{fig:chi_mlp} does indeed give us the ChI. For example, we could fuse the soft max outputs i.e., \emph{decision-in-decision-out} (DIDO) fusion of multiple deep learners (e.g., ResNet and GoogleNet). On the other hand, if ChIMP was pushed \emph{back} in the network, possibly connected directly to the inputs, it would likely function differently, e.g., \emph{signal-in-signal-out} (SISO) or SIDO versus DIDO. For example, each LCS neuron could represent a matched filter 
and the selection network would pick one result. 
We mention this because it (that is, context)
is substantial for XAI.
\end{remark}

\begin{figure} 
  \centering
    \includegraphics[width=0.36\textwidth]{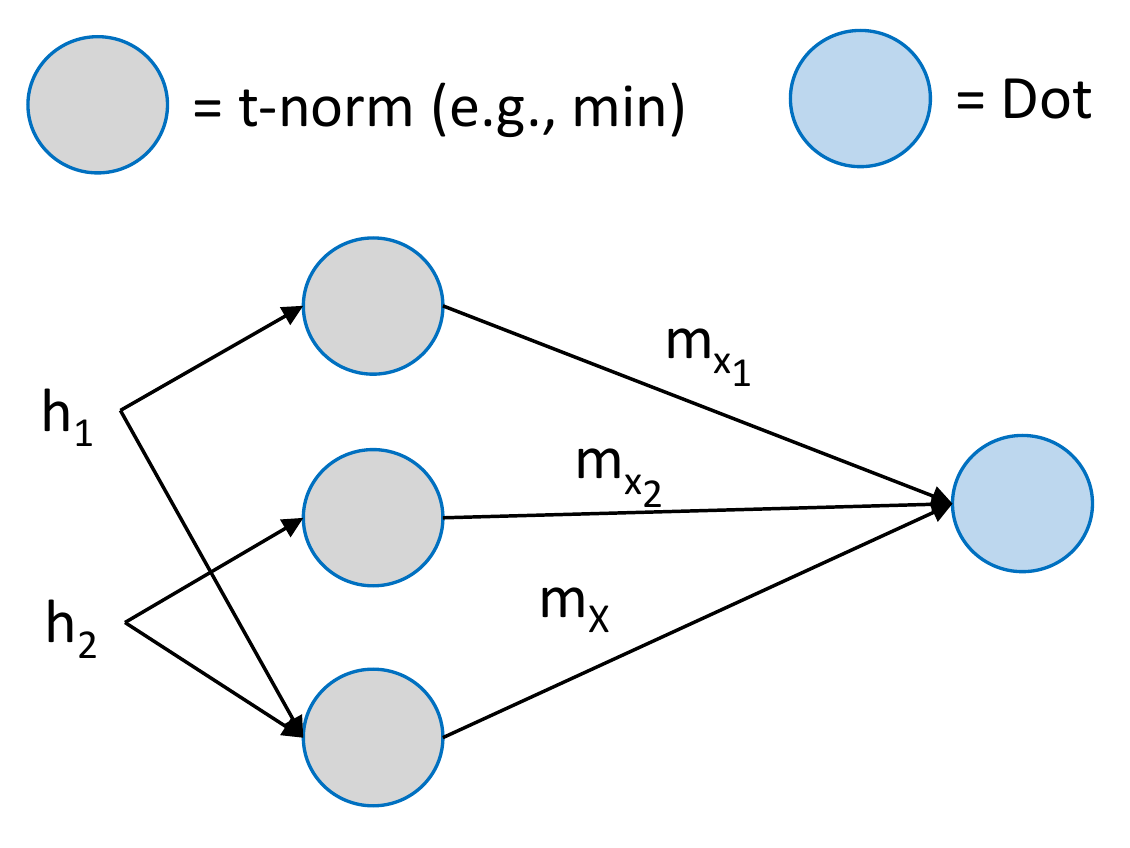}
  \caption{Mobius transform-based ChIMP.}\label{fig:chi_mob_mlp}
\end{figure}

\begin{remark}
Our $N!$ LCS-based ChIMP is not the only solution. Another example is based on the Mobius transform; see Fig.~\ref{fig:chi_mob_mlp}. The Mobius transform of $g$ is
\begin{align}
m(A) = \sum_{B\subseteq A} (-1)^{|A\setminus B|} g(B), \forall A \subseteq X,
\end{align}
which is invertable via the Zeta transform,
\begin{align}
g(A) = \sum_{B \subseteq A}{m(B)}, \forall A \subseteq X.
\end{align}
The Mobius transform of the ChI is
\begin{align}
C_g(\h) = \sum_{A \subseteq X} m(A) \bigwedge_{x_i \in A}{h_i}. 
\end{align}
Thus, the Mobius ChI can be thought of as a dot product of Mobius terms 
and a t-norm of the integrand term $\h$. 
There is no sort in the Mobius integral; the tradeoff is summing across $N!$ versus $N$ values. The point is, there are many ChIMPs. We presented two, but more of them likely exist. 
\end{remark}

\begin{remark}
The above $N!$ LCS-based and Mobius transform-based ChIMPs do not scale well with respect to $N$. For example, the Mobius-based ChIMP has $2^N$ t-norm neurons and one dot product. At higher levels (closer to the output, e.g., DIDO) in a neural network, $N$ might not be large (on the order of 10 classes) so all is well. However, if we consider using ChIMP at lower-levels in a network, a large $N$ could render ChIMP intractable. For example, consider fusing a set of convolutional filters of size $11 \times 11$. When unrolled, the $11 \times 11$ gives rise to $N=121$. As $2^{121}$ is a very large number, one could reduce ChIMP network complexity with respect to a method like k-additivity,
\begin{align}
C^k_g(\h) = \sum_{A \subseteq X, |A| \leq k} m(A) \bigwedge_{x_i \in A}{h_i}.
\end{align}
which uses tuples only up to, and including, $k$. The point is this, as $N$ grows the ChI can be restricted to suit the needs of an application at the expense of loss of some expressability.
\end{remark}

In summary, our response to Q1 is yes, an NN can represent the ChI and therefore a wide class of useful aggregation operators from the min to max, average, and more exotic variants as well. Furthermore, there are multiple ways (architectures) to achieve this. Technically, there are an infinite number of possibilities, e.g., recursive argument in which each solution is expanded by a single neuron, which could be bypassed or turned off by setting its weights to all zeros. This problem---existence of multiple ways to encode a solution---is well-known in many communities, e.g., bloating in genetic programming, which can be addressed using cost function augmentation with a complexity term. In the next section we explore an iChIMP architecture, which is simple to optimize using SGD and whose weights are explicit, enabling XAI.

\section{iChIMP}\label{sec:ichimp}

In this section we present iChIMP, an NN with an SGD solution. As such, this addresses Q2. To this end, we explore an alternative way of writing the ChI \cite{jin2018discrete},\footnote{Note that the ChI formulation herein differs from article \cite{jin2018discrete} in one respect that Eq.~\ref{eq:newchirep} is for $\mathbb{R}$-valued inputs whereas that in \cite{jin2018discrete} is for $\mathbb{R}^+$ inputs.} 
\begin{align}
C_g(\h) = 
\sum_{A \subseteq X} g(A) o(A),
\label{eq:newchirep}
\end{align}
where 
\begin{align}
    o(A) = 
    \begin{cases}
    \max \left( 0 , \bigwedge_{x_i \in A}{h_i} - \bigvee_{x_j \notin A}{h_j}\right), & A \subset X,\\
    \bigwedge_{x_i \in A}{h_i}, & A = X.
    \end{cases}
    \label{eq:ichmpp-oa}
\end{align}

\subsection{Measure Network}\label{sec:mnetwork}
\begin{figure} 
  \centering
    \includegraphics[width=0.48\textwidth]{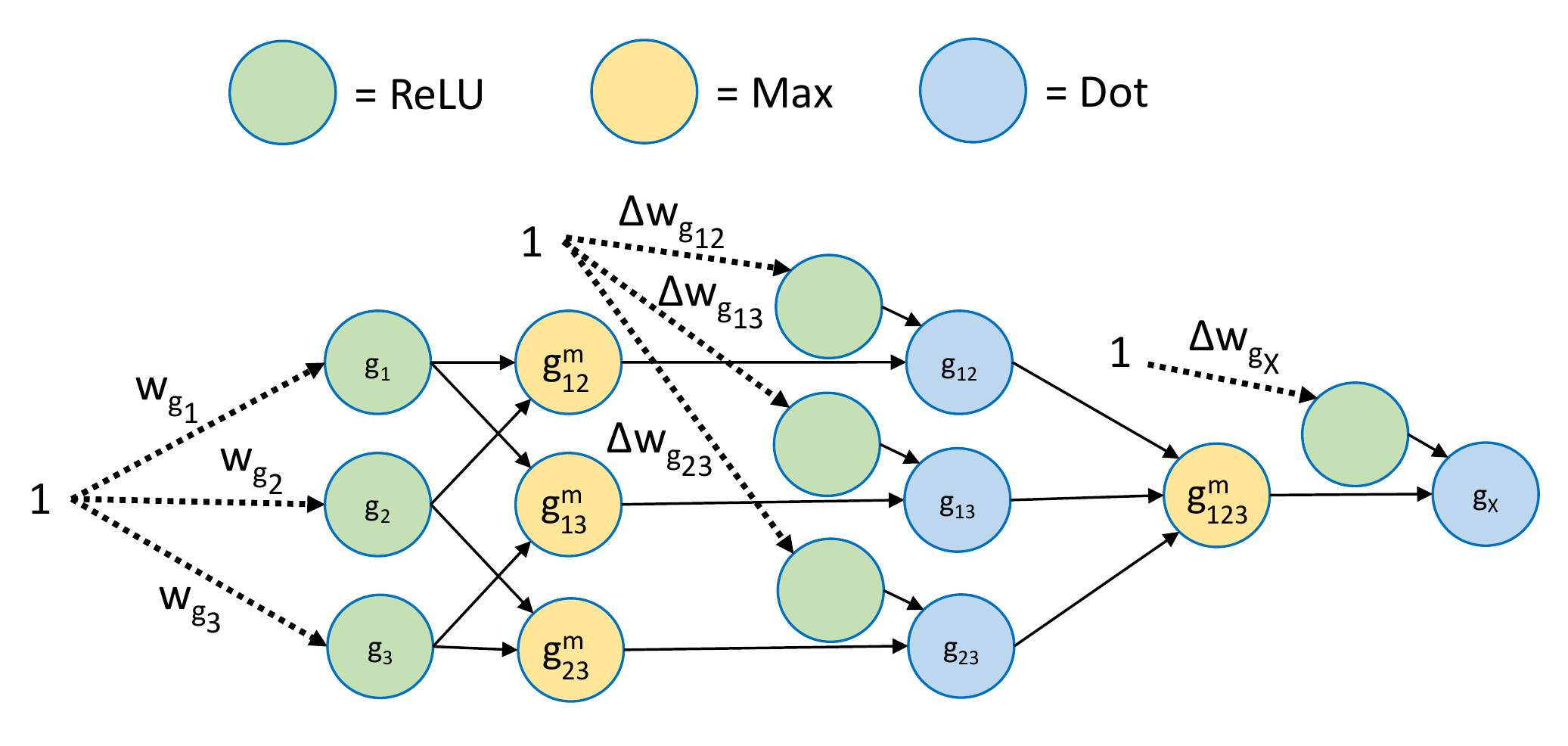} 
  \caption{FM learnable network for $N=3$. Dashed lines are learnable weights and dot has fixed $\w=(1,...,1)^t$.}\label{fig:f_g}
\end{figure}

Our idea is to design an FM NN. This network consists of constants, learnable weights, and existing neural mathematics (dot product, ReLU nonlinearlity, and maximum neurons). Figure \ref{fig:f_g} illustrates the network for $N=3$.

Specifically, the densities (FM variables whose set cardinality is 1) are represented as a weight vector and a nonlinearity (e.g., ReLU) enforces the lower boundary condition. Next, each tuple is expressed as
\begin{align}
g(A) = \bigvee_{B \subset A} g(B) + \Delta g(A),
    \label{eq:ichmpp-ga}
\end{align}
\noindent where $\Delta g(A) \in \Re$. Like before, a positive value enforcing nonlinearity is used to ensure the monotonicity property, forcing $\Delta g(A)$ to reside in $\Re^+$. If $g(X)$ is required to be $1$, then all of $g$ can be renormalized by taking the minimum of $g$ and $1$.
Otherwise, we can ignore the upper boundary condition, since this is not a hard requirement. 

\subsection{Integral and Evaluation Networks}\label{sec:intnetwork}
\begin{figure} 
  \centering
    \includegraphics[width=0.48\textwidth]{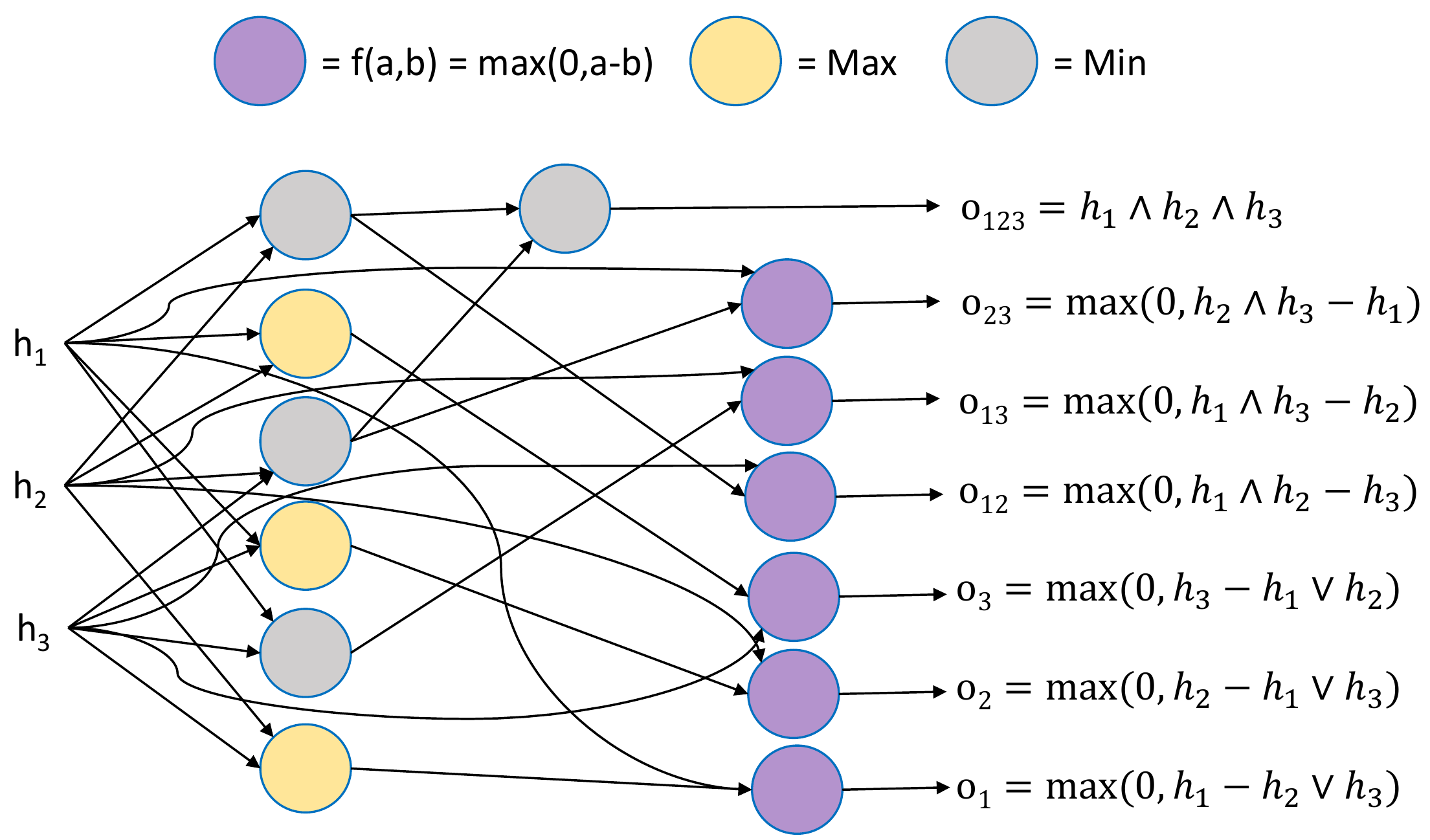} 
  \caption{Neural architecture for the integrand and $N=3$. Note, there are no learnable weights.}\label{fig:f_h}
\end{figure}

\begin{figure} 
  \centering
    \includegraphics[width=0.30\textwidth]{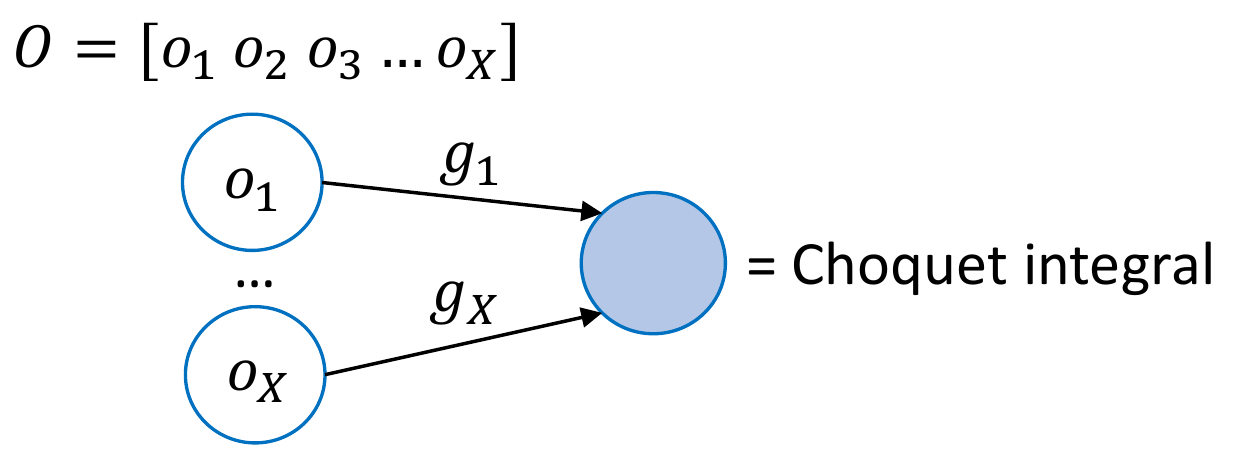} 
  \caption{Neural architecture that combines Fig.~\ref{fig:f_g} and \ref{fig:f_h}.}\label{fig:chi_combinestep}
\end{figure}

The next piece of iChIMP is expanding the $2^N-1$ terms in Eq.~\eqref{eq:newchirep} as shown in Fig.~\ref{fig:f_h}. This MLP has no trainable weights. The network can be achieved using max, min, and a custom $f(a,b)=max(0,a-b)$ neuron. The final step is a single dot product (see Fig.~\ref{fig:chi_combinestep}) of the expanded integrand terms (see Fig.~\ref{fig:f_h}) and the FM variable values (see Fig.~\ref{fig:f_g}).


\subsection{iChIMP Optimization} 
For readability, the derivation of iChIMP SGD optimization is presented in the Appendix.

\section{XAI for the Choquet Integral}\label{sec:inds}

In this section, a benefit of designing an explicit neural fusion network is highlighted. In \cite{bryce2018explainable}, we established ChI indices for XAI. The reader can refer to \cite{bryce2018explainable} for full mathematical explanation. Due to manuscript length, we are only able to summarize the indices.

The first category of fusion XAI indices explain the \emph{quality} of the individual sources and their interaction characteristics. The utility of a source, e.g., deep model, can be extracted via the Shapley index, $\Phi_{g}(i) \in [0,1]$, where $\sum_{i=1}^{N}{\Phi_{g}(i)}=1$. On the other hand, the Interaction Index \cite{murofushi1993techniques}, $\I_{g}(i,j) \in [-1,1]$, informs us about how two deep models \emph{interact} with one another---that is, is there an advantage in combining sources. A value of $1$ represents the maximum complementary between $i$ and $j$. On the other hand, a value of $-1$ represents the maximum redundancy between $i$ and $j$.

A second category of fusion XAI indices tell us what aggregation was learned. This helps us determine if the data are being combined in a union, intersection, average, or perhaps more unique and worthy of the ChI fashion. In \cite{bryce2018explainable}, we posed distance formulas to measure how similar a learned $g$ is to the maximum, minimum, mean, and in general, LOS.

The third category of fusion XAI indices is \emph{data centric}. In \cite{bryce2018explainable}, we determined how often an FM variable is encountered in training data; which helps us find missing FM variables. We also calculated what percentage of FM variables were observed in training data, what percentage of the $N!$ possible LCSs were observed, and if there is a dominant walk (and therefore lack of training data variety). These indices ultimately inform us about the quality of a learned solution and they highlight what is incomplete with respect to our model/training data. We also postulated a trust index based on what percentage of missing variables are used in a ChI calculation.

In summary, in \cite{bryce2018explainable} we discussed existing methods and proposed new ways to elicit information about a learned fusion. Since iChIMP is an explicit neural architecture, meaning we know which network elements map to which FM variables, XAI can help us understand, validate, and do iterative development.

\section{Experiments and Results}

In this subsection, two experiments are performed. The first experiment uses synthetic data. As such, we know the answer and we can control all factors, e.g., noise. We generate familiar operators that range from optimistic union-like to pessimistic intersection-like, average-like, and random operators. The purpose is to show range and variation in the FM and our ability to learn it. Next, we take the validated iChIMP and we use it to fuse a set of heterogeneous architecture DCNNs, to which we note no one knows the solution. The purpose of this experiment is to demonstrate the iChIMP on real-world benchmark data and to compare it to existing work.

\subsection{Experiment 1: Synthetic Data Set}
The objective of Experiment 1 is to show that we recover the correct ChI and to compare iChIMP to an existing (non-neural) way of solving the ChI, i.e., \emph{quadratic programming} (QP) \cite{Anderson2014}. Our data are generated pseudo-randomly from a uniform distribution in a unit interval and consist of $M=300$ samples and three inputs ($N=3$). We use four FMs with disparate aggregation behavior---FM1 is a soft-max, FM2 is a mean, FM3 is a soft-mean, and FM4 is an arbitrary FM---to generate their labels. Table \ref{tab:FMs} shows the four target FMs.

\begin{table}
\renewcommand{\arraystretch}{1.3}
  \centering
  \caption{Target FMs for Experiment 1.}
    \begin{tabular}{lccccccc}
    \toprule
    \textbf{FM}    & $g_1$  & $g_2$  & $g_3$  & $g_{12}$ & $g_{13}$ & $g_{23}$ & $g_{123}$ \\
    \midrule
    \textbf{FM1}   & 0.7   & 0.7   & 0.7   & 0.9   & 0.9   & 0.9   & 1.0 \\
    \textbf{FM2}   & $0.\bar{33}$   & $0.\bar{33}$   & $0.\bar{33}$   & $0.\bar{66}$   & $0.\bar{66}$   & $0.\bar{66}$   & 1.0 \\
    \textbf{FM3}   & 0.1   & 0.1   & 0.1   & 0.3   & 0.3   & 0.3   & 1.0 \\
    \textbf{FM4}   & 0.1   & 0.2   & 0.3   & 0.3   & 0.5   & 0.7   & 1.0 \\
    \bottomrule
    \end{tabular}%
  \label{tab:FMs}%
\end{table}%

In order to investigate the impact of noise on learning, we perturb the true labels with random noise sampled from a Gaussian distribution with variance $\sigma_y^2$. We consider six noise levels ranging from no-noise to $50\%$ of the true label standard deviation, i.e., $\sigma_n = \{0, 0.01\sigma_y, 0.05\sigma_y, 0.1\sigma_y, 0.3\sigma_y, 0.5\sigma_y\}$.

The data are partitioned into two segments, $80\%$ for training and $20\%$ for test. Training parameters for iChIMP are learning rate = $0.001$ and number of epochs = $1000$. The iChIMP variables are initialized with soft-mean like FM with values randomly picked from a uniform distribution in $[0.1,0.2]$. For each FM, optimization is performed on the training data to learn the FM variables, which are then used to estimate the label/output of the test data. The \emph{mean squared error} (MSE) with respect to the true training labels, true test labels, and FM variables are computed and used as a performance metric. The optimization task was repeated $20$ times for iChIMP with different initializations, and we report the average of the resultant MSEs. Table \ref{tab:results_noisyData} contains the results of Experiment 1.

\begin{table*}
  \centering
  \caption{Experiment 1 Label and Variable Error Rates at Noise Levels $\sigma_n = \{0, 0.01\sigma_y, 0.05\sigma_y, 0.1\sigma_y, 0.3\sigma_y, 0.5\sigma_y\}$.}
    \begin{tabular}{lcccccccccccc}
        \toprule
          & \multicolumn{6}{c}{Label Error}               & \multicolumn{6}{c}{FM Variable Error} \\
          \cmidrule(lr){2-7}\cmidrule(lr){8-13}
          & 0     & 0.01  & 0.05  & 0.1   & 0.3   & 0.5   & 0     & 0.01  & 0.05  & 0.1   & 0.3   & 0.5 \\
          \midrule
    FM1 ChIMP & 1.2E-15 & 3.4E-08 & 8.5E-07 & 3.4E-06 & 3.1E-05 & 8.5E-05 & 5.3E-15 & 1.4E-07 & 3.4E-06 & 1.4E-05 & 1.2E-04 & 3.4E-04 \\
    FM1 QP & 1.4E-05 & 1.4E-05 & 1.8E-05 & 2.2E-05 & 5.4E-05 & 1.1E-04 & 1.1E-04 & 1.1E-04 & 1.4E-04 & 1.7E-04 & 3.3E-04 & 6.1E-04 \\
    FM2 ChIMP & 1.2E-18 & 3.1E-08 & 7.9E-07 & 3.1E-06 & 2.8E-05 & 7.9E-05 & 9.5E-18 & 1.3E-07 & 3.1E-06 & 1.3E-05 & 1.1E-04 & 3.1E-04 \\
    FM2 QP & 5.1E-06 & 3.9E-06 & 3.7E-06 & 4.8E-06 & 2.4E-05 & 6.7E-05 & 3.4E-05 & 2.8E-05 & 2.2E-05 & 2.1E-05 & 7.1E-05 & 2.0E-04 \\
    FM3 ChIMP & 4.1E-20 & 3.4E-08 & 8.4E-07 & 3.4E-06 & 3.0E-05 & 8.4E-05 & 3.1E-19 & 1.3E-07 & 3.3E-06 & 1.3E-05 & 1.2E-04 & 3.3E-04 \\
    FM3 QP & 1.1E-05 & 1.1E-05 & 1.1E-05 & 1.4E-05 & 4.0E-05 & 8.9E-05 & 5.8E-05 & 5.9E-05 & 5.8E-05 & 6.8E-05 & 1.7E-04 & 3.5E-04 \\
    FM4 ChIMP & 1.8E-19 & 3.4E-08 & 8.4E-07 & 3.4E-06 & 3.0E-05 & 8.4E-05 & 1.1E-18 & 1.3E-07 & 3.4E-06 & 1.3E-05 & 1.2E-04 & 3.4E-04 \\
    FM4 QP & 2.8E-06 & 2.5E-06 & 3.1E-06 & 4.7E-06 & 2.7E-05 & 7.3E-05 & 1.6E-05 & 1.4E-05 & 1.5E-05 & 1.8E-05 & 8.8E-05 & 2.4E-04 \\
    \bottomrule
    \end{tabular}%
\label{tab:results_noisyData}
\end{table*}%

Table \ref{tab:results_noisyData} tells the following story. The MSEs for individual methods and their differences are quite low (on the order of $10^{-4}\sim 10^{-5}$) even at noise levels as high as $0.5\sigma_y$. This means that even though iChIMP is optimizing a non-convex network (with its ReLU, max, and min functions), it provides an approximation of the integrals as good as the QP method. 

\subsection{Experiment 2: Real-World Data Set}

Experiment 1 validates iChIMP and Experiment 2 uses it to fuse a set of heterogeneous \emph{deep CNNs} (DCNN) for remote sensing. An outstanding challenge in deep learning is network architecture. In general, no architecture has been shown to be superior across data sets. This is why we investigate the fusion of different architectures.

Two benchmark remote sensing data sets are investigated herein for land cover classification and object detection. The \emph{aerial image data set} (AID) has 30 classes, it has approximately 330 images per class, and the ground sampling distance (GSD) varies between 0.5 to 8 meters. The \emph{remote sensing imagery scene classification-45} (R45) was specifically designed to be challenging for remote sensing image scene classification. It contains 45 classes with 700 images per class and a variable GSD of 0.2 to 30 meters. However, the vast majority of the R45 classes have a GSD $<$ 1 meter.

Herein, we fuse seven DCNNs that have shown promising results in computer vision: DenseNet \cite{huang2017densely}, GoogLeNet \cite{Googlelenet_pub}, InceptionResNetV2 \cite{szegedy2017inception}, CaffeNet \cite{CaffeNet_pub}, ResNet-50 \cite{resnet}, Xception \cite{chollet2017xception}, and ResNet-101. For both data sets, our DCNNs were trained using the methods in \cite{CGINets}, including transfer learning (non-remote sensing weights derived from ImageNet), data augmentation (rotation, noise, scale, and contrast), and $50\%$ dropout. The trained DCNNs are then used in a locked state, i.e., no further learning happens during the fusion stage. The training of the DCNNs are done in five-fold, cross validation manner; we have 5 sets of 80\% training and 20\% testing for both data sets. 
Per DCNN fold, three-fold cross validation (CV) fusion is used (due to limited volume of data).
Table \ref{tab:exp2results} summarizes the performance of the DCNNs and our fused solution for the test data sets. In particular, iChIMP outperforms the individual networks; reducing the error rate  by 40\% (3.8\% down to 2.27\%) over the best single DCNN architecture for AID, and similarly a 30\% relative error rate reduction for R45.


\begin{table*}[]
\renewcommand{\arraystretch}{1.3}
  \centering
  \caption{Experiment 2 ChIMP Accuracy Results on Benchmark AID and R45 Data Sets.}
    \begin{tabular}{@{}ll*{8}{c}@{}}
    \toprule
          &       & \textbf{CaffeNet} & \textbf{DenseNet} & \textbf{GoogleNet} & \textbf{InceptionResNetV2} & \textbf{ResNet101} & \textbf{ResNet50} & \textbf{Xception} & \textbf{Shared ChIMP} \\
          \midrule
      \multirow{7}{*}{AID}  & Fold 1 & 93.55 & 95.40 & 95.70 & 96.20 & 96.20 & 95.65 & 97.40 & \textbf{97.80} \\
          & Fold 2 & 93.00 & 94.90 & 95.30 & 93.75 & 96.15 & 96.15 & 96.90 & \textbf{97.75} \\
     & Fold 3 & 94.40 & 94.35 & 94.80 & 95.35 & 95.30 & 95.20 & 95.70 & \textbf{96.95} \\
          & Fold 4 & 93.60 & 95.40 & 95.00 & 93.40 & 95.05 & 95.30 & 95.70 & \textbf{97.15} \\
          & Fold 5 & 94.70 & 94.65 & 94.80 & 95.95 & 96.15 & 96.10 & 96.80 & \textbf{97.40} \\
\cmidrule(lr){3-10}
& Mean  & 93.85 & 94.94 & 95.12 & 94.93 & 95.77 & 95.68 & 96.50 & \textbf{97.41} \\
          & SD   & 0.69  & 0.46  & 0.38  & 1.28  & 0.55  & 0.44  & 0.76  & \textbf{0.37} \\
          \midrule
      \multirow{7}{*}{R45}    & Fold 1 & 93.17 & 94.81 & 94.70 & 95.43 & 95.25 & 95.17 & 96.43 & \textbf{97.29} \\
          & Fold 2 & 93.41 & 95.59 & 95.60 & 95.65 & 95.60 & 95.44 & 95.97 & \textbf{97.19} \\
     & Fold 3 & 92.79 & 94.67 & 95.48 & 95.76 & 95.43 & 95.35 & 95.97 & \textbf{97.32} \\
          & Fold 4 & 93.51 & 93.60 & 95.62 & 95.51 & 95.68 & 95.41 & 96.11 & \textbf{97.59} \\
          & Fold 5 & 93.29 & 95.08 & 95.46 & 95.76 & 95.62 & 95.57 & 96.06 & \textbf{97.30} \\
          \cmidrule(lr){3-10}
          & Mean  & 93.23 & 94.75 & 95.37 & 95.62 & 95.52 & 95.39 & 96.11 & \textbf{97.34} \\
          & SD   & 0.28  & 0.73  & 0.38  & \textbf{0.15}  & 0.17  & \textbf{0.15}  & 0.19  &\textbf{ 0.15} \\
          \bottomrule
    \end{tabular}%
  \label{tab:exp2results}%
\end{table*}%


In most NNs, accuracy is the prime objective, and sometimes the only objective. However, we can apply our XAI indices to ``open up'' the learned solutions and interpret what is going on. On R45, the Shapley values are $0.1411 (\mu) \pm 0.0014 (\sigma), \	0.1414 \pm 0.0013, \	0.1418 \pm 0.0009, \	0.1415 \pm 0.0011, \	0.1437 \pm 0.0008,	\ 0.1431 \pm 0.0002,$ and $0.1474 \pm 0.0037$, with respect to CaffeNet, DenseNet, GoogLeNet, InceptionResNetV2, ResNet50, ResNet101, and Xception. On AID, the Shapley values are $0.1413 \pm 0.0010, \	0.1419 \pm 0.0003, \ 	0.1420 \pm 0.0011, \	0.1439 \pm 0.0007, \	0.1422 \pm 0.0005, \	0.1420 \pm 0.0007,$ and $0.1468 \pm 0.0028$. These Shapley values indicate that the deep nets have near equal overall importance, with perhaps the exception of Xception. Next, our XAI aggregation operator distance indices had a value of approximately $0$ relative to the mean. As such, we know that we learned an additive measure, which is reinforced by our Interaction Index values near $0$. At the moment, our XAI indices are \emph{evidence}. That is, their results need to be analyzed by an expert to determine what is going on. In future work we will discover a way to automatically reason about our various XAI information to deduce high-level linguistic summaries for non-experts.

In prior work \cite{scott2017fusion}, we explored the ``offline'' fusion---QP versus iChIMP---of three DCNNs. Herein, we repeat those experiments using iChIMP. This is done because we are interested to see if fusion learned a mean in part due to adding more deep models. As we discovered in \cite{scott2017fusion}, we do not always learn equal Shapley values. For example, on RSD we fused CaffeNet, GoogLeNet, and ResNet50. The shared weight Shapley solution had Shapley values of $0.28$, $0.4$, and $0.32$, respectively. Furthermore, if we trained a different iChIMP per class, versus a single shared weight solution across classes, then we obtain Shapley values of $0.06$, $0.69$, and $0.25$ for the bridge class and $0.04$, $0.32$, and $0.64$ for the mountain class. This informs us that while a single shared weight iChIMP has the best overall accuracy, using fewer deep models leads to non-uniform Shapley values and non-mean like aggregations. Furthermore, it informs us that different classes prefer different deep models. The question we need to address is why?

In \cite{bryce2018explainable}, we created XAI indices that tell us which FM variables cannot be approximated from training data. To no surprise, based on the above we ran our XAI indices and found that the ``problem classes'' that are bringing down the overall average per-class iChIMP solution are likely due to lack of training data variety (aka, they have a large percentage of un-approximated FM variables). This is probably a contributing factor to why the shared weight iChIMP performs better than $N$ per-class iChIMPs and possibly why a mean aggregation might be a good general strategy for combining a larger number of deep models; e.g., seven versus three, which means we need to encounter $7!=5{,}040$ versus $3!=6$ unique sorts.

In \cite{bryce2018explainable} we created an XAI index called dominant walk. A dominant walk is when a large percentage of the input data is sorted in a specific permutation order. We ran this index on our iChIMP solutions and discovered that there is typically a very dominant walk order of $h_1 \geq h_2 \geq ... \geq h_N$, the default order. This finding and pattern was too coincidental for our liking. Upon inspection, we discovered this is because a majority of our data has all the networks $100\%$ certainty in a label (and $0$s otherwise). As such, $1, 2,..., N$ is the default sort order. This means that we have strong learners and there is not a lot of diverse information (variety) to help us learn fusion. As such, a better solution, to be addressed in future work, would be to learn these networks in parallel now that the ChI can be integrated into a homogeneous neural solution.

In summary, in this subsection we used iChIMP to fuse a set of heterogeneous DNNs. Furthermore, iChIMP has XAI tools that allow us to understand, explain, and pursue the design of new data collections and better architectures.

\subsection{Computational Complexity}
In this subsection, a time complexity analysis is provided. This study consists of three steps, (i) assessing the complexity of $o(A)$, (ii) $g(A)$, and finally (iii) $C_g$, the dot the product of $g$ and $o$, as in Eq.~(\ref{eq:newchirep}).
\begin{enumerate}
    \item $o(A)$: For $A\subset X$, the computation of $o(A)$---see Eq.~(\ref{eq:ichmpp-oa})---has one minimum on $x$ ($1\le x \le N-1$) elements, one maximum on $N-x$ elements,  one maximum on two elements, and one subtraction. This gives $N+3$ operations as $O(n)$ is the worst case complexity for finding a maximum of $n$ elements and $O(1)$ for subtraction operations. Using a similar analysis, the number of operations for $A=X$ is $N$. As a result, the total cost for computing $o(A)$ for all $A$, where $A\subseteq X \text{ and } A \neq \emptyset$, is $(2^N-2) \times (N+3) + N$. 
    The complexity of $o(A)$ is the highest term without the constant, i.e., $O(N2^N)$.
    \item $g(A)$: Eq. (\ref{eq:ichmpp-ga}) for $g(A)$ has two parts: (a) $\Delta g(A)$, a Relu on two elements (thus $O(2)$ complexity); and (b) $\bigvee_{B \subset A} g(B)$, a maximum on $|A|-1$ elements with cost of $|A|-1$, where $|A|$. As a result, the cost of $g(A)$ is $|A|+1$. Let $\prescript{N}{}{C_{x}}$ be the combination of $N$ elements taken $x$ time. Thus, there are $\prescript{N}{}{C_{|A|}}$ sets with cardinality $|A|$.
    The total cost of computing $g(A)$ for all $A$ is 
    \[\prescript{N}{}{C_{1}} \times (1+1) + \dots +  \prescript{N}{}{C_{N}} \times (N+1) < 2^N (N+1),\]
because    $\prescript{N}{}{C_{0}} + \dots +  \prescript{N}{}{C_{N}}  = 2^N$. This results in a time complexity of $O(N2^N)$ for computation of $g(A)$.
    \item $C_g$: Eq. (\ref{eq:newchirep}) involves the dot product of $g$ and $o$ with $2N$ \textit{floating point operations} (FLOPS) including $N$ additions and $N$ multiplications, resulting in a $O(N)$ complexity. 
\end{enumerate}

Combining all three components, the time complexity of iCHIMP is $O(N2^N) + O(N2^N) + O(N) = $ $O(N2^N)$. This means that iChIMP has complexity of exponential order, $O(N2^N)$ w.r.t. the number of inputs, and complexity of polynomial order, $O(Mlog(M))$, with respect to the number of FM variables, $M = 2^N$.

Next, we discuss the cost of fusing deep models in practical applications. First, many pre-trained models, e.g., the ones used herein for large datasets like Imagenet, are publicly available. Common practice is to apply transfer learning, which is an offline procedure that is inexpensive relative to training an equivalent network from scratch. Next, the fusion of $N$ models only adds $2^N$ variables, where $N$ is minuscule in relation to the number of parameters in the deep models. Translated, offline learning of $N$ fusion variables is nominal. Online is a similar story. iChIMP is a tiny parallel network, relative to the complexity of a deep model, of common deep learning operations that have been accelerated in tool kits like TensorFlow and PyTorch. The only worthwhile cost of fusing $N$ deep models is the storage and computational cost associated with evaluating $N$ deep models, which can be performed in parallel. However, as evident in our experiments, fusion improves performance. Overall, the expense of fusing a set of deep models can be rationalized relative to the current dilemma of not knowing which deep model is correct.

\section{Conclusion and Future Work}

In this article, we proposed a novel NN architecture with a gradient descent-based optimization solution that mimics the Choquet integral for information aggregation. This solution was demonstrated on synthetic data for validation and real-world benchmark data sets in remote-sensing fusion relative to the fusion of a set of heterogeneous architecture deep models. Adding iChIMP to the top of a set of deep neural networks produces a deep fuzzy neural network. Furthermore, we analyzed similarities and differences between multi-layer nets and the ChI, with respect to factors like representation and constrained learning algorithms. Last, our recently established XAI indices were used to ``open up'' our learned deep neural solutions enabling interpretability, helping us understand what was learned, identifying flaws in our training data, and ultimately designing better deep model solutions. 

The proposed NN architectures are not just limited to scalar-valued inputs and can be applied to higher-order FSs, such as Type-1 or Type-2, using the Zadeh’s extension principle.  If one wishes to use a NN for higher-order FS-valued inputs, and can define an objective function and associated gradients, then using our proposed ChI-based NN is as straightforward as it is for real-valued inputs. Extensions of the fuzzy ChI for FS-valued inputs have been previously proposed \cite{6722924}; we suggest application and extension of iChIMP for follow-on work.

In future work, now that we have the iChIMP foundation we will explore efficient representations---e.g., $k$-additivity or our recently established data-compressed ChI \cite{islam2017data}. We will also investigate advanced learning algorithms---e.g., drop out \cite{srivastava2014dropout}, regularization \cite{goodfellow2016deep}, etc.---with regard to the $(2^N-2)$ free variables. Once this is achieved we can push the ChI neuron back in the network for signal- and feature-level fusion, versus decision-level fusion. Furthermore, we will explore where and when a fusion neuron should be used, akin to the current revolution of what architecture should be employed. We will also, in the future, explore the possible benefit of enforcing $g(X)=1$, focusing on what exact penalty functions to use to enforce a soft boundary. We will also continue to explore XAI and discover ways to linguistically summarize their contents for human consumption and to use them possibly in optimization directly to encourage certain factors, e.g., diversity, specificity, or efficiency. In this paper, we used iChIMP to fuse pre-trained DCNNs. Next, we will explore how to simultaneously learn iChIMP and the component networks to improve factors like accuracy and network diversity.

\newpage

\appendix

\subsection{Derivative of maximum}\label{appendix:max}

The derivative of $f = \max\{f_1(x), f_2(x)\}$ is 
\begin{equation}
\frac{df(x)}{dx} = 
\begin{cases}
\frac{df_1(x)}{dx}  & \quad \text{ if } f_1(x) > f_2(x)\\
\frac{df_2(x)}{dx}  & \quad \text{ if } f_1(x) < f_2(x)\\
\frac{1}{2} \left(\frac{df_1(x)}{dx} + \frac{df_2(x)}{dx}\right)  & \quad \text{ if } f_1(x) = f_2(x).\\
\end{cases}
\label{eq:maxDerivative}
\end{equation}
Let $J_{f=f_i}$ be an indicator function that points to whether $f_i$ is equal to $f$ (in other words, max of $f_i$s) or not, i.e.,
\begin{align*}
    J_{f=f_i} =
    \begin{cases}
    1 & \quad \text{ if }  f(x) = f_i(x) \\
0 & \quad \text{ else.}
\end{cases}
\end{align*}
As such, we can write \eqref{eq:maxDerivative} as
\begin{align}
\frac{df(x)}{dx} = \frac{\sum_{i=1}^2 J_{f=f_i} \frac{df_i(x)}{dx}}{\sum_{i=1}^2 J_{f=f_i}} = \sum_{i=1}^2 I_{f=f_i} \frac{df_i(x)}{dx},
    \label{eq:derivative-max}
\end{align}
where $I_{f=f_i} = \frac{J_{f=f_i}}{\sum_{i=1}^2 J_{f=f_i}}$ is a normalized indicator variable. Respectively, \eqref{eq:derivative-max} can be generalized for the case of $f(x)=$ $\max \{f_1(x),\dots, f_n(x)\}$ as  
\[\frac{df(x)}{dx} =  \sum_{i=1}^n I_{f=f_i} \frac{df_i(x)}{dx}.\]

\subsection{Derivative of min}\label{appendix:min}
The derivative for $f(x) = \min \{f_1(x), $ $\dots, f_n(x)\}$ is
\[
\frac{df(x)}{dx} = \sum_{i=1}^n I_{f=f_i} \frac{df_i(x)}{dx}.
\]

\subsection{Gradients of weights}\label{appendix:weights}
Let $y_k$, $k =1,\dots, M$, be the iChiMP output for observation $o_k$. For a set of data, the error is
\begin{align}
E = \frac{1}{2} \sum_{k}^{M}{ (l_k - y_k)^{2} }.  
\end{align}
For notational simplicity, we define $2^N-1-N$ (not defined on the densities nor empty set) max of subset auxiliary variables (see Fig.~\ref{fig:f_g}), $g^m$. For example,
\begin{align*}
g^m_{123}= & \max{(g_{12},g_{13},g_{23})},\\
g_{123}= & g^m_{123}+\max{(\Delta w_{g_{123}},0)}.
\end{align*}
Without loss of generality, the following intermediate formulas are provided relative to $N=3$,
\begin{align}
        \frac{\partial E}{\partial y_k} & = (y_k - l_k) = e_k,\\       
        \frac{\partial y_k}{\partial g_i} & = o_i,\\
        \frac{\partial g_{123}}{\partial g^m_{123}} & = 1,\\
\frac{\partial g^m_{123}}{\partial g_{12}} & = I_{(g^m_{123} = g_{12})} + 0 + 0 = I_{(g^m_{123} = g_{12})}.
\end{align}
The same holds for $\frac{\partial g^m_{123}}{\partial g_{13}}$ and $\frac{\partial g^m_{123}}{\partial g_{23}}$ respectively. 
As such,
\begin{align}
        \frac{\partial g_{123}}{\partial g_{12}} & = \frac{\partial g_{123}}{\partial g^m_{123}}\frac{\partial g^m_{123}}{\partial g_{12}} =  I_{g^m_{123} = g_{12}},
\end{align}
and
\begin{align}
\frac{\partial g_{123}}{\partial \Delta w_{g_{123}}} = \left\{ \begin{array}{cc} 
                1 & \hspace{5mm} \Delta w_{g_{123}} > 0 \\
                0 & \hspace{5mm} \text{else}. \\
                \end{array} \right.
\end{align}
Furthermore, we define the indicator variable 
\[I_{c>0} = 
\begin{cases}
1 & \quad \text{ if } c>0\\
0 & \quad \text{ else,}
\end{cases}
\] 
and thus
\[\frac{\partial g_{123}}{\partial \Delta w_{g_{123}}} = I_{\Delta w_{g_{123}}>0}.\]
Next, the gradients are
\begin{align}
        \frac{\partial E}{\partial g_{123}} & = \frac{\partial E}{\partial y_{k}}\frac{\partial y_k}{\partial g_{123}} = e_k o_{123},
\end{align}
\begin{subequations}
\begin{align}
        \frac{\partial E}{\partial g_{12}} & = \frac{\partial E}{\partial y_k}\left( \frac{\partial y_k}{\partial g_{12}} + \frac{\partial y_k}{\partial g_{123}} \frac{\partial g_{123}}{\partial g_{12}} \right) \\ 
        &=e_k \left(o_{12}+o_{123} I_{g^m_{123} = g_{12}} \right), 
\end{align}
\end{subequations}
and similar for $\frac{\partial E}{\partial g_{13}}$ and $\frac{\partial E}{\partial g_{23}}$. In a similar fashion, the error for each density is defined as
\begin{subequations}
\begin{align}
        \frac{\partial E}{\partial g_{1}} = & \frac{\partial E}{\partial y_k} \bigg( \frac{\partial y_k}{\partial g_{1}} + \frac{\partial y_k}{\partial g_{12}}\frac{\partial g_{12}}{\partial g_{1}}+  \\ 
        & \frac{\partial y_k}{\partial g_{13}}\frac{\partial g_{13}}{\partial g_{1}} + \frac{\partial y_k}{\partial g_{123}}\frac{\partial g_{123}}{\partial g_{1}} \bigg)  \\
        = & e_k\big( o_1 + o_{12} I_{g_{12}^m=g_1} + o_{13} I_{g_{13}^m=g_1}+ \\
        & o_{123} \left(I_{g_{123}^m=g_{12}} I_{g_{12}^m=g_1} + I_{g_{123}^m=g_{13}} I_{g_{13}^m=g_1}\right)\big),
\end{align}
\end{subequations}
and similar for $\frac{\partial E}{\partial g_{2}}$ and $\frac{\partial E}{\partial g_{3}}$. Last, 
\begin{subequations}
\begin{align}
        \frac{\partial E}{\partial \Delta w_{g_{123}}} & = \frac{\partial E}{\partial g_{123}}\frac{\partial g_{123}}{\partial \Delta w_{g_{123}}} \\
        & = e_k o_{123}I_{\Delta w_{g_{123}}>0},
\end{align}
\end{subequations}
\begin{subequations}
\begin{align}
        \frac{\partial E}{\partial \Delta w_{g_{12}}} & = \frac{\partial E}{\partial g_{12}}\frac{\partial g_{12}}{\partial \Delta w_{g_{12}}} \\
        & = e_k (o_{12} + o_{123} I_{g^m_{123} = g_{12}})I_{\Delta w_{g_{12}}>0},
\end{align}
\end{subequations}
and similar for $\frac{\partial E}{\partial \Delta w_{g_{13}}}$ and $\frac{\partial E}{\partial \Delta w_{g_{23}}}$.

\subsection{Gradients for inputs}\label{appendix:inputs}

Here we give the expressions for the gradients of the cost function with respect to inputs $h_1$, $h_2$, and $h_3$. 
The gradients of $o_i$s w.r.t.~$h_1$ are
\begin{align*}
\frac{\partial o_1}{\partial h_1} & = 0 + I_{o_1 = h_1 - h_2 \vee h_3}  = I_{o_1 = h_1 - h_2 \vee h_3},\\
\frac{\partial o_{2}}{\partial h_1} & = 0 + I_{o_2 = h_2 - h_1 \vee h_3} (0 - (I_{h_{13}^{max}=h_1}+0)) \\
 & = -I_{o_2 = h_2 - h_1 \vee h_3}I_{h_{13}^{max}=h_1},\\
\frac{\partial o_{3}}{\partial h_1} &  =  -I_{o_3 = h_3 - h_1 \vee h_2}I_{h_{12}^{max}=h_1},\\
\frac{\partial o_{12}}{\partial h_1} & = \left(0 + I_{o_{12} = h_1\wedge h_2 - h_3 }  (I_{h_{12}^{min} = h_1}+0)\right) \\
 & =  I_{o_{12} = h_1\wedge h_2 - h_3 }I_{h_{12}^{min} = h_1},\\
\frac{\partial o_{13}}{\partial h_1} & = \left(0 + I_{o_{13} = h_1\wedge h_3 - h_2}  (I_{h_{13}^{min} = h_1} +0)\right) \\
& =  I_{o_{13} = h_1\wedge h_3 - h_2}I_{h_{13}^{min} = h_1},\\
\frac{\partial o_{23}}{\partial h_1} & = 0 -I_{o_{23} = h_2\wedge h_3 - h_1} = -I_{o_{23} = h_2\wedge h_3 - h_1},\\
\frac{\partial o_{123}}{\partial h_1} & = I_{o_{123}=h_1} + 0 + 0 = I_{o_{123}=h_1}.
\end{align*}

The gradient of $h_1$ w.r.t.~the cost function, $E$, is
\begin{align*}
    \frac{\partial E}{\partial h_1} & = \frac{\partial E}{\partial y_k} \sum_i \frac{\partial y_k}{\partial o_i} \frac{\partial o_i}{\partial h_1} \\
    & = e_k \big(g_1 I_{o_1 = h_1 - h_2 \vee h_3} - g_2 I_{o_2 = h_2 - h_1 \vee h_3}I_{h_{13}^{max}=h_1} \\
    & - g_3 I_{o_3 = h_3 - h_1 \vee h_2}I_{h_{12}^{max}=h_1} +
    g_{12} I_{o_{12} = h_1\wedge h_2 - h_3 }I_{h_{12}^{min} = h_1} \\
    & + g_{13} I_{o_{13} = h_1\wedge h_3 - h_2}I_{h_{13}^{min} = h_1} - g_{23}I_{o_{23} = h_2\wedge h_3 - h_1} \\
    & + g_{123}I_{o_{123}=h_1}\big).
\end{align*}
Similarly, 
\begin{align*}
    \frac{\partial E}{\partial h_2} & = \frac{\partial E}{\partial y_k} \sum_i \frac{\partial y_k}{\partial o_i} \frac{\partial o_i}{\partial h_2} \\
    & = e_k \left( - g_1 I_{o_1 = h_1 - h_2 \vee h_3} I_{h_{23}^{max}=h_2} + g_2 I_{o_2 = h_2 - h_1 \vee h_3} \right.\\
    & - g_3 I_{o_3 = h_3 - h_1 \vee h_2}I_{h_{12}^{max}=h_2} +
    g_{12} I_{o_{12} = h_1\wedge h_2 - h_3 }I_{h_{12}^{min} = h_2} \\
    & - g_{13} I_{o_{13} = h_1\wedge h_3 - h_2} + g_{23}I_{o_{23} = h_2\wedge h_3 - h_1} I_{h_{23}^{min} = h_2} \\
    & \left. + g_{123}I_{o_{123}=h_2}\right),\\
    \frac{\partial E}{\partial h_3} & = \frac{\partial E}{\partial y_k} \sum_i \frac{\partial y_k}{\partial o_i} \frac{\partial o_i}{\partial h_3} \\
    & = e_k \left( - g_1 I_{o_1 = h_1 - h_2 \vee h_3} I_{h_{23}^{max}=h_3} \right.\\
    & - g_2 I_{o_2 = h_2 - h_1 \vee h_3} I_{h_{13}^{max}=h_3} \\
    & + g_3 I_{o_3 = h_3 - h_1 \vee h_2} -
    g_{12} I_{o_{12} = h_1\wedge h_2 - h_3 } \\
    & + g_{13} I_{o_{13} = h_1\wedge h_3 - h_2} I_{h_{13}^{min} = h_3} + g_{23}I_{o_{23} = h_2\wedge h_3 - h_1} I_{h_{23}^{min} = h_3} \\
    & \left. + g_{123}I_{o_{123}=h_3}\right).
\end{align*}

\bibliographystyle{IEEEtran}
\bibliography{Refs}


\end{document}